\def\year{2019}\relax
\documentclass[letterpaper]{article} 
\usepackage{aaai19}  
\usepackage{times}  
\usepackage{helvet}  
\usepackage{courier}  
\usepackage{url}  
\usepackage{graphicx}  
\usepackage{mathtools}
\usepackage{blindtext}
\usepackage{breqn}
\usepackage[utf8]{inputenc} 
\usepackage[T1]{fontenc}    
\usepackage[]{hyperref}       
\usepackage{url}            
\usepackage{booktabs}       
\usepackage{amsfonts}       
\usepackage{nicefrac}       
\usepackage{microtype}      
\usepackage{amsmath}
\usepackage{comment}
\usepackage[colorinlistoftodos]{todonotes}
\usepackage{bbm}
\usepackage{algorithm}
\usepackage[noend]{algpseudocode}
\usepackage{multirow}
\usepackage[normalem]{ulem}

\usepackage{changepage}
\usepackage{xcolor}
\DeclareUnicodeCharacter{2212}{-}

\newcommand\sys{{\textsc{TraPSNet}}}
\newcommand\sysshort{{\textsc{TNet}}}
\newcommand\torpido{{\textsc{ToRPIDo }}}

\begin{document}
\title{Size Independent Neural Transfer for RDDL Planning}
\author{Sankalp Garg, Aniket Bajpai, Mausam\\
Indian Institute of Technology, Delhi\\
New Delhi, India\\
{\tt \{sankalp2621998, quantum.computing96\}@gmail.com, mausam@cse.iitd.ac.in}\\
}
\maketitle
\begin{abstract}
Neural planners for RDDL MDPs produce deep reactive policies in an offline fashion. These scale well with large domains, but are sample inefficient and time-consuming to train from scratch for each new problem. To mitigate this, recent work has studied neural transfer learning, so that a generic planner trained on other problems of the same domain can rapidly transfer to a new problem. However, this approach only transfers across problems of the same size. We present the first method for neural transfer of RDDL MDPs that can transfer across problems of different sizes.  Our architecture has two key innovations to achieve size independence: (1) a state encoder, which outputs a fixed length state embedding by max pooling over varying number of object embeddings, (2) a single parameter-tied action decoder that projects object embeddings into action probabilities for the final policy. On the three challenging RDDL domains of SysAdmin, Game Of Life and Academic Advising, our approach powerfully transfers across problem sizes and has superior learning curves over training from scratch.

\end{abstract}

\section{Introduction}
Recently, deep reactive policies have been shown to be successful for offline probabilistic planning problems represented in RDDL \cite{fern18} or PPDDL \cite{asnet18}. An advantage is that neural policy networks can represent offline policies for very large domains while at the same time being competitive with traditional planners on small domains. However, training these networks from scratch can be sample-inefficient and time-consuming. 

Since neural planners learn latent representations, they present an opportunity to transfer their policy learned over one problem instance to other problem instances. This can be especially useful in the case of neural policy networks if the transfer time is much less than the training time. Very recently, we have proposed \torpido -- \textcolor{blue}{a}  neural transfer learning approach \cite{torpido}, which trains a generic policy network from some problems of an RDDL domain, and transfers it to a new problem of the same domain. 
\torpido\ learns a transformation from the state and action spaces to \emph{latent} state and action spaces, and then learns a neural policy in these latent spaces. At test time, for a new problem instance, it relearns a mapping from the latent action space to actual actions, and transfers the other components of the architecture. While highly effective in reducing learning times, a significant limitation of \torpido\ is that it only transfers when all training and testing problems are of the same size. Thus, it works only when test problem sizes are known at train time, which limits its applicability.

In response, we present \sys, the first {\em size-independent} neural transfer algorithm for RDDL MDPs. Furthermore, \sys\ also achieves significantly better transfer results, both in terms of reward and time in the same problem settings as \torpido\. As a first step towards this goal, our paper focuses on domains where action templates and (non-)fluents are parameterized over a single object only, and there is one binary non-fluent. A majority of domains from the International Probabilistic Planning Competition (IPPC) 2014 \cite{ippc14} satisfy these constraints.

\torpido can only operate on equi-sized problems, because its state embeddings have dimensionality proportional to the number of objects in the problem, and its action decoder outputs a distribution over all possible actions, whose number also depends on the problem size. 
\sys\ achieves size-independence through the use of two key ideas. First, it uses max pooling of object embeddings to produce a fixed-dimensionality state embedding. Max-pooling, intuitively, helps the state embedding retain the best value for each feature (dimension), while losing information about the specific object(s) responsible for that value. 
Second, while it still produces a probability distribution over all actions, it does so by projecting an object embedding onto the probability with which the action applied on that object is taken in the policy.  The parameters for this projection function are tied across all objects, making this size-independent also.

We perform experiments on three RDDL domains -- SysAdmin, Game of Life and Academic Advising. These are chosen because while they are highly challenging (because they can have very large state and transition spaces, and complex dynamics), they also satisfy the assumption of unary actions and binary non-fluents. Training on small-sized problems in these domains, and testing on larger instances in IPPC 2014, we find that \sys\ achieves excellent zero-shot transfer, i.e., it has a very high reward even before any RL on test problem. Compared to when training from scratch, \sys\ has vastly superior learning curves. We release the code of \sys\ for future research.\footnote{Available at {\url{https://github.com/dair-iitd/trapsnet}}}

\section{Related Work}
\label{background}
Most current state of the art reinforcement learners are neural models. 
A popular deep RL agent is Asynchronous Advantage Actor-Critic (A3C) \cite{a3c}, which simultaneously trains a policy and a value network, by running simulated trajectories and backpropagating an advantage function (which is a function of obtained rewards). We refer to the parameters in these two networks as $\theta_\pi$ and $\theta_V$, respectively. 

Any RL agent is naturally applicable to probabilistic planning, since any RDDL \cite{rddl} or PPDDL \cite{ppddl} problem can always be converted to a simulator for training the agent.  Existing literature on neural planning includes: Value Iteration Networks, which operate in flat state spaces \cite{vin},  Action-Schema Networks (ASNets) for solving PPDDL problems \cite{asnet18}, and deep reactive policies for RDDL problems \cite{fern18}. Some works have also studied neural transfer, which include Groshev et al. \shortcite{abbeel18}, which experiment on only two  deterministic domains, and ASNets. While all RDDL problems can, in principle, be converted to PPDDL, the potential exponential blowup of the representation makes ASNets unscalable to our domains. The closest to our work is \torpido -- our recent architecture for equi-sized transfer in RDDL domains \cite{torpido}.

\torpido\ is based on the principle that there may exist a latent embedding space for a domain, where similar states in different problems will have similar embeddings. It tries to uncover this latent structure using object connectivities exposed in RDDL via non-fluents. 
It has a {\em state encoder}, which creates object embeddings as projections of this graph adjacency matrix and fluents in a state (using a GCN), and then concatenates them to construct a (latent) state embedding. An {\em RL module} maps this state embedding to a (latent) state-action embedding.  An {\em action decoder} maps state-action embedding to a policy (distribution over action symbols). 
While other modules transfer directly, the decoder needs to be re-trained at test time, resulting in a {\em near} zero-shot transfer. 
Because its embeddings and action decoder are size-specific, \torpido\ only allows equi-sized transfers -- a limitation we relax in our work. Furthermore, \sys\ requires no retraining at start, and achieves a {\em full} zero-shot transfer.

\section{Problem Formulation}
An RDDL \cite{rddl} domain enumerates the various fluent predicates ($fl_j$), non-fluent predicates ($nf_l$), a reward function $R$ and action templates ($act_k$) with their dynamics. Here, fluents refers to predicates that can change value as a consequence of actions, whereas non-fluents stay fixed throughout execution -- these often describe the connectivity structure among objects in a problem.  An RDDL domain can be likened to a Relational Markov Decision Process  \cite{relationalmdp}. An RDDL problem within a domain lists the specific objects, values of all non-fluent predicates for those objects and the fluent values for those objects (which define state variables) in the initial state. This completes the description of a factored MDP with a known initial state \cite{mausam&kolobov12,kolobov-uai12}. Note that different problems can have differing sizes based on the number of objects. 

Our goal is to develop a good {\em anytime} algorithm for computing an offline policy $\pi_T$ for an RDDL test problem $P_T$. An anytime MDP algorithm is one that can be stopped at any time and will return a reasonable policy; it typically produces better policies given more computation time.  We use a transfer setting for this, where we are given $N$ training problems $P_1, P_2,\ldots,P_N$ from the same domain, but of different (typically smaller) size as that of $P_T$. The transfer objective is to, at training time, learn domain-specific but problem-independent information from training problems, and, at test time, transfer that to $P_T$.  

Post transfer, training further on $P_T$ should achieve a good anytime MDP planner. That has two indicators. First, in zero-shot setting, i.e., when the algorithm is given no access to $P_T$ simulator and cannot retrain, it must return a policy with a high long term reward. Second, it must have superior learning curves compared to a policy learned from scratch on $P_T$.

As the first step towards the objective of size-independent transfer, we focus on domains where all fluents, non-fluents and action templates are unary, except one non-fluent is binary. This is a common setting in many benchmark RDDL domains such as SysAdmin, Game Of Life and Academic Advising. Let our factored MDP have a set of objects $o\in O$, parameterized fluents $fl_j(o)$ and non-fluents $nf_l(o)$, actions $act_k(o)$, and a special parameterized binary non-fluent $nf(o,o')$.

\section{The \sys\ Architecture} \label{architecture}


We name our neural transfer model, \sys\ -- a \textbf{Net}work that can \textbf{Tr}ansfer \textbf{a}cross \textbf{P}roblem \textbf{S}izes. At a high level, it extends A3C and trains problem-independent policy and value networks. The transfer itself is based on two hypotheses: (1) for every domain, neural embeddings can capture similarities of objects and states across problem sizes; (2) Q-value of a specific action (say $act_k(o)$) can be effectively estimated via a problem-oblivious function that depends on the current state and $o$'s embeddings.

\begin{figure*}[t]
\centering
+\includegraphics[width=\linewidth]{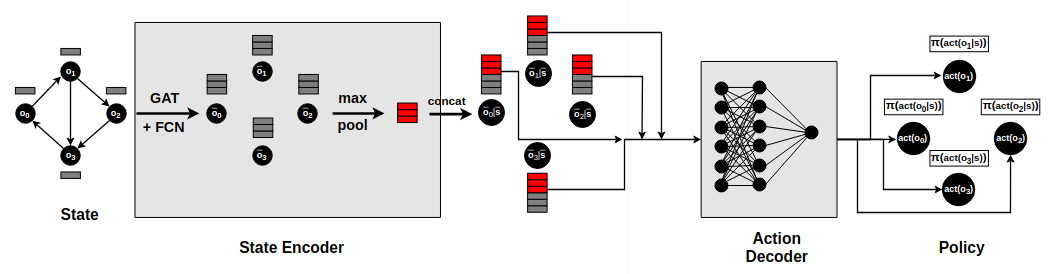}

\caption{\small Model architecture for \sys\ (only policy network is shown)}
\label{modelarch}

\end{figure*}

Both the policy and value net of \sys\ have a state encoder each, whose output feeds into an action decoder (for policy net), and a value decoder (for value net). The parameters of these modules are shared across all training problems in a domain. The state encoders operationalize the first hypothesis by outputting a fixed-size object embedding $\overline{o}$ for each object $o$ in the problem based on the non-fluent graph structure, and fluent values related to $o$. For different problems, a variable number of object embeddings are max-pooled to construct a fixed-size state embedding $\overline{s}$. The action decoder operationalizes the second hypothesis. It projects each object embedding $\overline{o}$, in conjunction with action id $k$ and the overall state embedding $\overline{s}$, to a real-valued score.  This is the un-normalized probability of taking $act_k(o)$ in the state $s$. All actions in a problem are run through a softmax to compute a randomized policy $\pi$. A similar idea is used in value decoder for estimating $V(s)$. Figure \ref{modelarch} illustrates the policy net of \sys\ schematically.

\subsection{State Encoder}
We want to construct an embedding for each object based on its individual properties, its neighborhood, and also the global information of the overall state. Similar to Bajpai et al. \shortcite{torpido}, this is achieved by casting the state information in a graph. The nodes of the graph are objects $o\in O$. There is an edge between $o$ and $o'$ if $nf(o,o')=1$. The input features at each node $o$ are the concatenated values of $fl_j(o)$ and $nf_l(o)$ . To compute fixed-size object embeddings $\overline{o}$, \sys\ constructs local embeddings for each node $o$ in the graph. For this purpose, it uses a Graph Attention Network (GAT)\footnote{Lack of space precludes a detailed description of a GAT \cite{gat}. Briefly, a GAT improves on a GCN by computing, in each node, self-attention coefficients for each neighbor and itself. These coefficients multiplied with node features and added to obtain an intermediate node embedding. This process is repeated K times and the results are max pooled to obtain a final node embedding.}
 followed by a fully connected layer. This takes in the adjacency matrix of the graph and outputs node embeddings.

It then computes an embedding $\overline{s}$ for the whole state $s$, i.e., the entire graph. To achieve a size-invariant $\overline{s}$, \sys\ pools all $\overline{o_i}$s to get $\overline{s}$. After experimenting with various pooling schemes (max, sum, average), max pooling produced the best results. Similar results have been seen in NLP and vision literature (e.g., \cite{maxpooling}). Max-pooling, intuitively, helps the state embedding retain the ``best" value for each feature (dimension), while losing information about the object(s) responsible for that value. For each object $o$, $\overline{s}$ is concatenated with $\overline{o}$ to produce a {\em contextual} object embedding, $\overline{[o|s]}$, which is used as input to both the decoders.



\subsection{Action \& Value Decoders}
A fully-connected network maps a  contextual $\overline{[o|s]}$ embedding into several real-valued scores for each object, one for each action template. Let these networks represent functions $f^\pi_k$ and $f^V$ for the action and value decoder, respectively. $f^\pi_k(\overline{[o|s]})$ is interpreted as a score for action $act_k(o)$ -- a softmax over $f^\pi_k$ for all $(o,k)$ pairs produces a randomized policy $\pi$. The value of a state $V(s)$ is approximated by value net as $\sum_{o\in O} f^V(\overline{[o|s]})$.

This architecture enables \sys\ to apply the same action decoder for problems of different sizes, since the network itself is not size-dependent --  it is replicated $|O|$ times (with tied parameters) to compute the values of each action. It also enables estimation of $V(s)$ in different ranges for problems of different sizes, akin to the sum of values of objects approximation in Relational MDPs \cite{guestrin-rmdp}. 



\subsection{Learning and Transfer}
\sys\ is trained end to end using a standard RL objective on $P_1,\ldots,P_N$. For each problem, its RDDL simulator interacts with the agent to generate trajectories. The rewards (advantage) obtained in these trajectories are backpropagated through value and policy nets according to the A3C loss to train $\theta_\pi$ and $\theta_V$. We make one small modification in which at each step the gradients are accumulated from trajectories of all training problems, so that the learned parameters do not overfit on any one problem.

At transfer time, pre-trained \sys\ can be run directly on $P_T$ using its adjacency matrix, to obtain an initial $\pi_T$, without any modification or retraining, since there are no problem-specific parameters. Due to this, we expect the model to have good zero-shot transfer performance.  Training using RL on $P_T$ improves the policy further.

\begin{figure*}[t]
\centering
\includegraphics[scale=0.3]{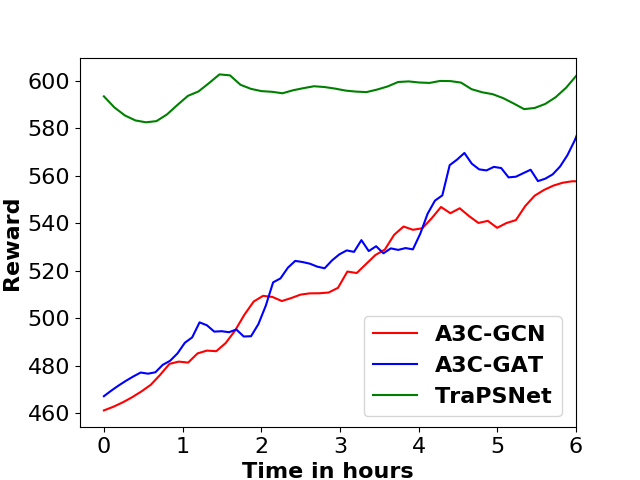}
\includegraphics[scale=0.3]{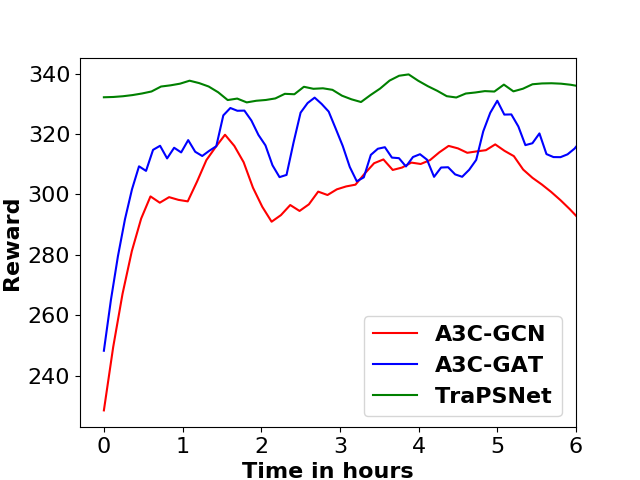}
\includegraphics[scale=0.3]{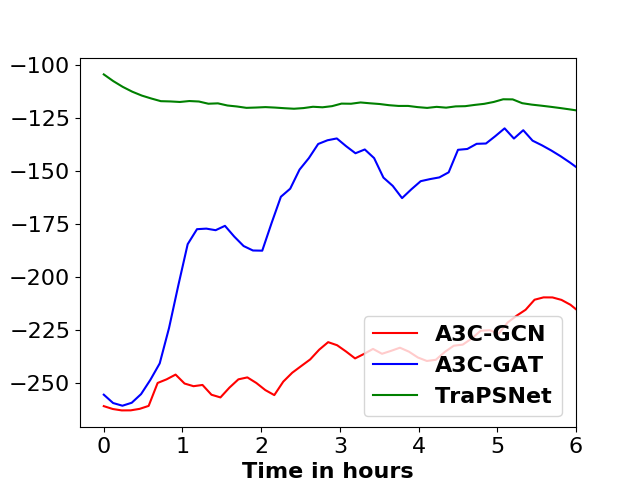}
\includegraphics[scale=0.3]{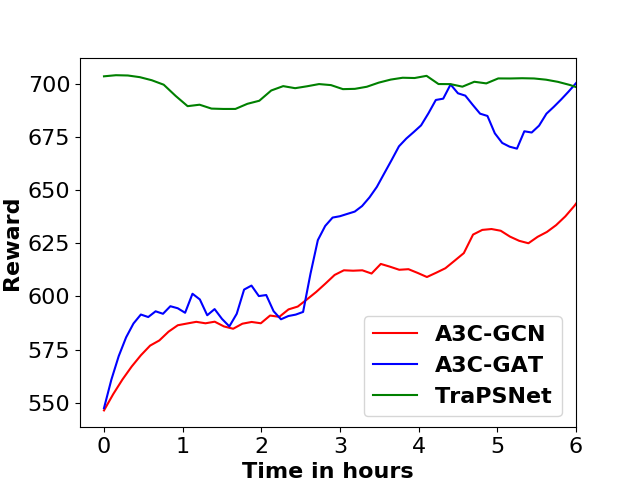}
\includegraphics[scale=0.3]{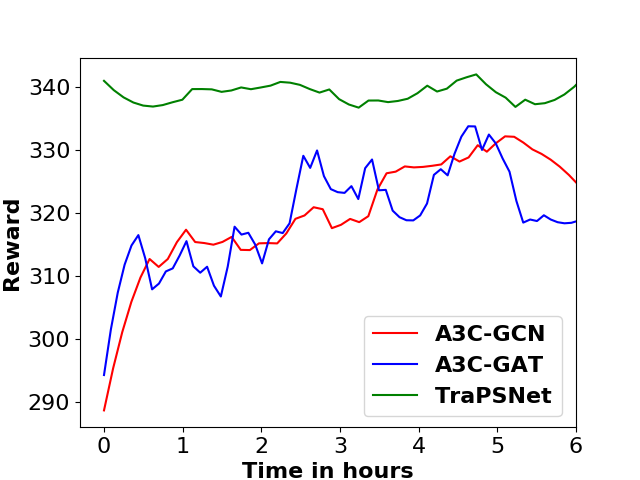}
\includegraphics[scale=0.3]{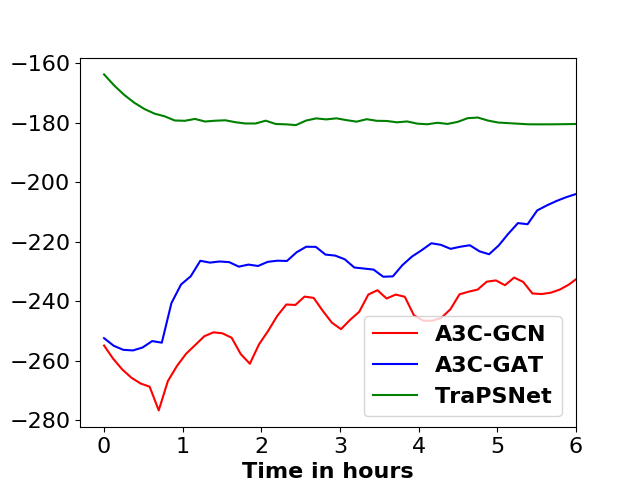}
\caption{Learning curves on $7^{th}$ and $9^{th}$ problems in IPPC 2014 for SysAdmin, Game Of Life and Academic Advising. \sys\ outperforms other baselines by wide margins -- reward is very high right from the beginning because of an effective zero-shot transfer. The graphs in columns represent sysadmin, game of life and academic advising respectively.}
\label{comparison_figure}
\end{figure*}

\begin{table*}[t]
\centering
\begin{tabular}{|c|c|c|c|c|c|c|c|c|c|}
\hline
Time (hrs) & \multicolumn{3}{c|}{0} & \multicolumn{3}{c|}{3} & \multicolumn{3}{c|}{6}\\ \hline
Arch. & A3C-GCN & A3C-GAT & \sysshort{} & A3C-GCN & A3C-GAT & \sysshort{} & A3C-GCN & A3C-GAT & \sysshort{}\\ \hline
Sys 5    & 0.00   & 0.07  & \textbf{0.84}  &  0.32   & 0.53   & \textbf{0.88}   & 0.54    & 0.79   & \textbf{0.89}   \\ 
Sys 6    & 0.06   & 0.04  & \textbf{0.63}  &  0.50   & 0.58   & \textbf{0.73}   & 0.65    & \textbf{0.87}   & 0.78   \\ 
Sys 7    & 0.03   & 0.04  & \textbf{0.92}  &  0.41   & 0.51   & \textbf{0.89}   & 0.69    & 0.78   & \textbf{1.00}   \\ 
Sys 8    & 0.00   & 0.06  & \textbf{0.93}  &  0.28   & 0.58   & \textbf{0.86}   & 0.51    & 0.53   & \textbf{0.89}   \\ 
Sys 9    & 0.00  & 0.06  & \textbf{0.89}  &  0.41   & 0.51   & \textbf{0.89}   & 0.57    & 0.87   & \textbf{0.92}   \\ 
Sys 10   & 0.05   & 0.1  & \textbf{0.88}  & 0.25    & 0.50   & \textbf{0.93}   & 0.31    & 0.52   & \textbf{0.92}   \\ \hline
GoL 5  & 0.00   & 0.07  & \textbf{0.83}  & 0.48    & 0.68   & \textbf{0.87}   & 0.71    & 0.79   & \textbf{0.85}    \\ 
GoL 6  & 0.00   & 0.06  & \textbf{0.77}  & 0.62    & 0.61   & \textbf{0.88}   & 0.56    & 0.58   & \textbf{0.88}    \\ 
GoL 7  & 0.00   & 0.05  & \textbf{0.88}  & 0.71    & 0.69   & \textbf{0.92}   & 0.60    & \textbf{0.90}   & 0.88    \\ 
GoL 8  & 0.00   & 0.10  & \textbf{0.70}  & 0.74    & 0.75   & \textbf{0.86}   & 0.71    & \textbf{0.96}   & 0.78    \\ 
GoL 9  & 0.00   & 0.08  & \textbf{0.90}  & 0.56    & 0.78   & \textbf{0.87}   & 0.75    & 0.48   & \textbf{0.93}    \\ 
GoL 10 & 0.05   & 0.10  & \textbf{0.32}  & 0.78    & \textbf{0.84}   & 0.28   & 0.79    & \textbf{1.00}   & 0.35    \\ \hline
Acad 5  & 0.07   & 0.00  & \textbf{0.93}  & 0.61    & 0.93   & \textbf{0.98}   & 0.86    & 0.91   & \textbf{0.94}    \\ 
Acad 6  & 0.02   & 0.01  & \textbf{0.99}  & 0.00    & 0.94   & \textbf{0.99}   & 0.31    & 0.95   & \textbf{0.97}    \\ 
Acad 7  & 0.27   & 0.01  & \textbf{1.00}  & 0.30    & 0.80   & \textbf{0.94}   & 0.33    & 0.78   & \textbf{0.92}    \\ 
Acad 8  & 0.39   & 0.03  & \textbf{0.92}  & 0.18    & 0.76   & \textbf{0.86}   & 0.61    & \textbf{0.97}   & 0.90    \\ 
Acad 9  & 0.00   & 0.05  & \textbf{0.99}  & 0.44    & 0.53   & \textbf{0.93}   & 0.45    & 0.54   & \textbf{0.90}    \\ 
Acad 10 & 0.20   & 0.01  & \textbf{0.90}  & 0.30    & 0.35   & \textbf{0.93}   & 0.44    & 0.65   & \textbf{0.94}    \\ \hline
\end{tabular}
\caption{Comparison of $\alpha(t)$ values of \sys\ (\sysshort) against baselines A3C-GCN and A3C-GAT at three different training points of 0, 3 and 6 hours. \sys\ outperforms or obtains comparable performance for almost all problems.}
\label{comparison_table}

\end{table*}

\section{Experiments} 
\label{experiments}
Our experiments evaluate the ability of \sys\ to perform zero-shot transfer, as well as compare its anytime performance to training from scratch.


\subsection{Domains}
We use three RDDL benchmark domains from International Probabilistic Planning Competition 2014: SysAdmin \cite{sysadmin}, Game of Life (GoL) \cite{rddl} and Academic Advising \cite{acad}. These are chosen, because they are challenging due to their large state spaces and complex dynamics, but also amenable for our algorithm because their  non-fluent is binary and actions unary. Briefly, each SysAdmin problem has a network of computers (arranged in different topologies via non-fluent {\em connected}), and the goal is to keep as many computers {\em on} as possible. The agent can {\em reboot} a computer in each step. Each Game of Life problem represents a grid world (of a different size). Each cell is alive or dead, and the agent can make one cell alive in each time step. The goal is to keep as many cells alive as possible. Each Academic Advising problem represents a student in a university trying to graduate by completing his degree requirements. The student needs to pay a certain cost to register for a course (can be different for different courses). Courses can be compulsory or optional. The probability of passing a course depends on the number of pre-requisite courses completed by the student. The student is also charged a fixed cost for each semester in the university. The goal is for the student to complete his degree at minimum cost.

\subsection{Experimental Settings}
For each domain, we train \sys\ on randomly generated problem instances of small sizes and then test on benchmark problems of larger sizes. For SysAdmin, we use $N = 5$ training problems with 10, 11, 12, 13 and 14 computers, and test on IPPC problems 5 to 10, which have 30 to 50 computers. 
For Game of Life, we use $N = 3$ different problems with 9 cells each, and again test on IPPC problems 5 to 10, which have 16 to 30 cells. For Academic Advising, we use $N = 2$ different problems with 10 courses each, and again test on IPPC problems 5 to 10, which have 21 to 30 courses. The largest test problems are SysAdmin 9 and 10, with state space of $2^{50}$, and 50 available actions. 

We use the same hyperparameters for all problem instances of all domains keeping in the spirit of domain independent planning. The GAT layer of state encoders uses a neighbourhood of 1. It takes in one feature per node as input, and outputs 3 features per node. A fully connected layer then projects this into a 20-dimensional space for $\overline{o}$, which is also the dimensionality of $\overline{s}$.  The action and value decoders are 2-layer fully connected networks with an intermediate layer of size 20. All layers use a leaky ReLU activation as non-linearity. \sys\ is trained using RMSProp with a learning rate of $10^{−3}$. All models are written in TensorFlow and run on an Ubuntu 16.04 machine with Nvidia K40 GPUs.



\subsection{Baselines \& Evaluation Metrics} 
To the best of our knowledge, no size-invariant transfer algorithm exists for RDDL domains. We compare against our base non-transfer algorithm, A3C. For fairness, we augment A3C with GCN (which is already known to outperform A3C \cite{torpido}). We also compare against A3C-GAT, to verify if the benefit is due to GAT or transfer.

We note that the results of neural architectures are not directly comparable to modern symbolic planners like PROST \cite{prost}. This is because neural architectures are offline planners, i.e., after the training is accomplished, the policy can be run with little computation (in this case, one forward pass). However, PROST is an online planner -- it has a significant deliberation phase after observing the result of each action. This difference makes the two planners incomparable.

We measure the transfer capability of our model using the evaluation metrics from Bajpai et al. \shortcite{torpido}. We measure the performance of our model at intermediate training times $t$ by simulating the policy network at $t$ upto the specified execution horizon, and averaging the values. This simulation is run 100 times to get a stable result. We call this value $V_{\pi} (t)$. We report $\alpha (t) = (V_{\pi} (t) - V_{inf}) / (V_{sup} - V_{inf}) $, where $V_{sup}$ and $V_{inf}$ represent the highest and lowest values obtained on the current problem by any planning algorithm at any $t$. $\alpha (t)$ signifies the fraction of best performance achieved at time $t$. Moreover,  $\alpha (0)$ acts as a measure of zero-shot transfer.


\subsection{Results} 
Figure \ref{comparison_figure} compares the training curves of \sys\ and the two baselines  for the $7^{th}$ and $9^{th}$ problems of both domains (first column is for SysAdmin, the next is for game of life and last is for academic advising). The curves plot $V_{\pi}(t)$ as a function of training time $t$. Comparing the baselines, we notice that a GAT improves the performance over a GCN for SysAdmin. 
We also observe that \sys\ demonstrates excellent zero-shot transfer, obtaining a very high initial reward. It is vastly superior to the baselines, because before training they can only act randomly. As training times increase, \sys's anytime performance remains better or very close to the baselines for most problems. In many cases, baselines after 6 hours cannot even match up to \sys's performance at the start.  This underscores the importance of our transfer algorithm.

The detailed $\alpha(t)$ results for all problems (at three training times) are reported in Table \ref{comparison_table}, which corroborates these observations. An exception is GoL problem 10, where after excellent initial transfer, \sys's performance does not match up to that of the other baselines. Further investigation reveals that all training (and other test) problems in the domain are square grids, whereas problem 10 is the only rectangular grid (10$\times$3). We suspect that the training has overfit somehow on the squareness of the grid. 

 
\section{Conclusions}
We present \sys, the first neural transfer algorithm for RDDL MDPs that can train on small problems of a domain and transfer to a larger one. This requires \sys\ to maintain a size-invariant ;atent representation of the state, which is achieved by pooling over object embeddings, and the use of a parameter-tied action decoder, which projects objects onto corresponding actions. Experiments show vastly superior performance compared to training from scratch. 

Our work brings the classical formulation of Relational MDPs back to the fore. We believe neural latent spaces may overcome the limitations of a traditional sum of symbolic basis function representation used previously for this problem \cite{sanner-uai05}. While we demonstrate results for a specific kind of Relational MDPs, in future, we plan to study the robustness and generality of this approach for other types of RDDL domains.

\section*{Acknowledgements}
This work is supported by grants from Google, Bloomberg, IBM and 1MG, and a Visvesvaraya faculty award by Govt. of India. We thank Microsoft Azure sponsorships, and the IIT Delhi HPC facility for computational resources.

\bibliography{icaps19}
\bibliographystyle{aaai}

\end{document}